\documentclass{article}

\usepackage[normalem]{ulem}

\usepackage[final,nonatbib]{neurips_2025_diffsys}

\usepackage[utf8]{inputenc} 
\usepackage[T1]{fontenc}    
\usepackage{hyperref}       
\usepackage{url}            
\usepackage{booktabs}       
\usepackage{amsfonts}       
\usepackage{nicefrac}       
\usepackage{microtype}      
\usepackage{xcolor}         
\usepackage{graphicx}       
\usepackage{amsmath}
\usepackage[style=numeric-comp,sorting=none]{biblatex}
\usepackage{todonotes}
\usepackage{wrapfig} 
\usepackage{cleveref}
\crefname{figure}{Fig.}{Figs.}
\crefname{table}{Table}{Tables}
\crefname{section}{Sec.}{Secs.}
\crefname{equation}{Eq.}{Eqs.}
\usepackage[inline]{enumitem}

\title{
 WARP-LUTs: Walsh-Assisted Relaxation for Probabilistic Look Up Tables
}

\author{%
  Lino Gerlach \\
  Princeton University\\
  \texttt{lg0508@princeton.edu} \\
  \And
  Liv Våge \\
  Princeton University \\
  \texttt{lv7805@princeton.edu} \\  
  \And
  Thore Gerlach \\
  University of Bonn \\
  \texttt{tgerlac1@uni-bonn.de} \\
  \And
  Elliott Kauffman \\
  Princeton University \\
  \texttt{ek8842@princeton.edu} \\
  \And
  Isobel Ojalvo \\
  Princeton University \\
  \texttt{iojalvo@princeton.edu} \\
}

\begin{document}

\maketitle

\begin{abstract}
Fast machine learning is of growing interest to the scientific community and has spurred significant research into novel model architectures and hardware-aware design. Recent hard- and software co-design approaches have demonstrated impressive results with entirely multiplication-free models, such as Differentiable Logic Gate Networks~(DLGNs). However, these models suffer from high computational cost during training and do not generalize well to logic blocks with more inputs.
In this work, we introduce Walsh-Assisted Relaxation for Probabilistic Look-Up Tables~(WARP-LUTs)---a novel gradient-based method that efficiently learns combinations of logic gates with substantially fewer trainable parameters. We demonstrate that WARP-LUTs achieve significantly faster convergence on CIFAR-10 compared to DLGNs, while maintaining comparable accuracy. Furthermore, our approach suggests potential for extension to higher-input logic blocks, motivating future research on extremely efficient deployment on modern FPGAs and its real-time science applications.



\end{abstract}

\section{Introduction}
\label{sec:introduction}


\begin{wrapfigure}{r}{0.42\textwidth} 
    \vspace{-20pt} 
    \centering
    \includegraphics[width=0.45\textwidth]{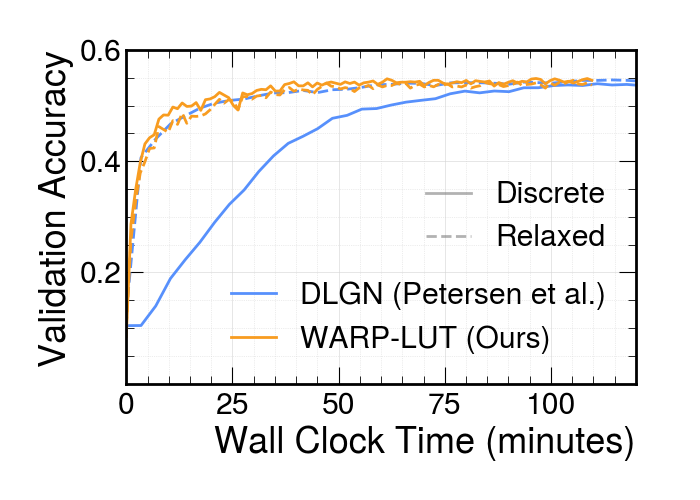}
    \vspace*{-23pt}
    \caption{Validation accuracy on CIFAR-10. Logic gate activations are represented as continuous numbers (relaxed) during training and discrete values during inference. Our implementation shows faster convergence compared to the baseline.}
    \label{fig:moneyplot}
    \vspace{-10pt} 
\end{wrapfigure}

Deep learning~\cite{lecun2015deep} has become the standard for a wide range of tasks in science,
but its success comes with heavy computational costs in both training and inference. This restricts deployability in many real-world settings---particularly in domains where ultra-fast inference is critical, such as particle physics~\cite{Aarrestad_2021}, gravitational wave astronomy~\cite{martins2025improvingearlydetectiongravitational}, and quantum computing~\cite{bhat2024machinelearningarbitrarysinglequbit}.
These constraints have motivated substantial research into the development of models that maintain predictive accuracy while improving computational efficiency.
Prominent among such efforts are model compression techniques, which encompass approaches for inducing sparsity~\cite{sung2021training,hoefler2021sparsity}, pruning~\cite{lin2018accelerating,liu2022unreasonable}, and reducing numerical precision via parameter quantization~\cite{sun2024gradient,gholami2022survey,chmielneural}.

Nevertheless, such approaches do not directly address the intrinsic computational cost of numerical multiplication.
To overcome this limitation, multiplication-free architectures have been proposed, including binary neural networks~\cite{hubara2016binarized,qin2020binary} and other bit-level summation models~\cite{chen2020addernet,elhoushi2021deepshift,nguyen2024bold}.
However, efficient inference on logic gate–based hardware requires mapping such abstract computations into executable logic, which incurs significant overhead.

Within the domain of multiplication-free models, Weightless Neural Networks~(WNNs)~\cite{aleksander2009brief} represent a distinct class that overcomes this limitation.
Instead of relying on weighted connections, WNNs employ look-up tables~(LUTs) with binary values to drive neural activity during inference, allowing them to capture highly nonlinear behaviors while avoiding arithmetic operations altogether.
Although state-of-the-art models achieve remarkable performance on small-scale tasks, their scalability remains constrained---whether due to limited expressiveness~\cite{susskind2023uleen}, reliance on gradient approximations~\cite{bacellar2024differentiable}, double-exponential growth in parameter requirements~\cite{petersen2022deep,petersen2024convolutional}, or large discretization gaps~\cite{kim2023deep,yousefi2025mind}, 
describing an accuracy mismatch between the training using relaxations and discrete inference.

We overcome these issues by proposing \textbf{\underline{W}alsh-\underline{A}ssisted \underline{R}elaxation for \underline{P}robabilistic \underline{LUTs}~(WARP-LUTs)}, based on differentiable relaxation similar to Differentiable Logic Gate Networks~(DLGN)~\cite{petersen2022deep}.
Publishing our code will bridge the gap created by the lack of publicly available convolutional DLGN implementations for the community~\cite{yousefi2025mind}.
Our contributions can be summarized as follows:
\begin{itemize}
    \item We propose representing Boolean functions with the Walsh--Hadamard~(WH) transform~\cite{kunz1979equivalence}, providing a compact and differentiable parameterization of a deep WNN enabling an exponential reduction in the number of parameters compared to DLGNs.
    \item Through single-exponential parameter scaling in LUT-input size, this not only increases expressiveness~\cite{carneiro2019exact}, but also maps directly to modern FPGA primitives, such as LUT-$6$ blocks, promising more efficient deployment on such hardware.
    \item A Gumbel reparameterization scheme is presented. This yields a smoother loss landscape, better alignment between training and inference, faster convergence, and a reduced discretization gap.
    \item Experiments show the versatility and efficacy of our approach compared to previous methods.
\end{itemize}

\section{Related Work}
\label{sec:related_work}

Early work on WNNs explored single-layer architectures that replace weighted connections with LUTs, enabling extremely efficient inference~\cite{susskind2022weightless,susskind2023uleen}.
However, their limited expressiveness and lack of effective training methods for deeper architectures have restricted their use in complex tasks.

To address these limitations, DLGNs represent Boolean functions as differentiable two-input gates, enabling gradient-based training of multi-layer architectures~\cite{petersen2022deep,petersen2024convolutional}.
Advances in connection learning~\cite{mommen2025method,yue2024learning,kresse2025scalable} have improved flexibility, but training DLGN architectures introduces a discretization gap between continuous training and discrete inference.
Recent work has sought to bridge this gap through stochastic relaxation~\cite{kim2023deep} and forward-backward alignment~\cite{yousefi2025mind}.

However, DLGNs remain restricted to two-input gates and suffer from double-exponential parameter growth, limiting scalability.
To account for higher-degree LUTs, the authors of~\cite{bacellar2024differentiable} extend LUT-based models with differentiable training, but rely on gradient approximations that can affect performance.
Finally, alternative approaches move beyond neural networks, such as TreeLUT~\cite{khataei2025treelut}, which combines gradient-boosted trees with LUT mappings for efficient inference, and Truth Table Net~\cite{benamira2024truth}, which leverages Boolean circuit structures for scalability and verifiability.

\section{Methodology}
\label{sec:methodology}


Any Boolean function $f:\{0,1\}^n \to \{0,1\}$ is uniquely represented by its truth-table. Since the input- and output space have size $2^n$ and 2, respectively, there are $2^{(2^n)}$ boolean functions. Each one can be uniquely represented in the WH basis by first re-encoding its inputs from $\{0,1\}$ into signed variables $\{-1,+1\}$, according to the mapping $B:\{0,1\} \to \{-1,+1\}$, $B(0)=-1$, $B(1)=+1$.
The resulting function takes the form
\begin{align}
f(\mathbf x) = \mathrm{sign}\left(l_{\mathbf c,B}(\mathbf x) \right),\quad
l_{\mathbf c,B}(\mathbf x)=\sum_{S \subseteq \{1,\dots,n\}} c_S \prod_{i \in S} B(x_i),
\label{eq:walsh}
\end{align}
with $B(x_i) \in \{-1,+1\}$ describing the transformed inputs. The $2^n$-dimensional vector of all WH coefficients can be calculated from the orthogonal transform $\mathbf c = \frac{1}{2^n}Hf_{\pm} \in \tfrac{1}{2^n}\mathbb{Z}^{2^n}$,
where $f_{\pm}\in \{\pm1\}^{2^n}$ is the $\pm1-$valued truth vector and $H$ is the $2^n \times 2^n$ Hadamard matrix.
Intuitively, this expansion uses simple polynomial basis functions—individual variables, pairwise products, and higher-order interactions—to capture the structure of the Boolean function. Hence the WH representation provides a compressed and structured parametrization of Boolean functions, where the $2^{2^n}$ Boolean functions are in bijection with a finite subset of this $2^n$-dimensional lattice.



\paragraph{Example: 2-input logic gates}
In the special case of two inputs $(a,b)$, every binary logic gate admits a decomposition with only four coefficients:
\begin{align*}
f(a,b) = \mathrm{sign}\!\big(c_0 + c_1 a + c_2 b + c_3 (a \cdot b)\big),
\end{align*}
where \(c_0\) encodes the constant bias (tendency toward 0 or 1), \(c_1\) and \(c_2\) encode dependence on the individual inputs, and \(c_3\) encodes the interaction term between the inputs.
For example, the coefficients $(c_0,c_1,c_2,c_3)=(0,0,0,-1)$ correspond to the \textsc{XOR} gate, while the \textsc{AND} gate can be expressed as $(c_0,c_1,c_2,c_3)=(-\tfrac{1}{2},\tfrac{1}{2},\tfrac{1}{2},\tfrac{1}{2})$. This decomposition demonstrates that instead of enumerating all 16 binary gates explicitly, one can parameterize them compactly with just four WH coefficients (see~\cref{tab:binary-gates} in~\cref{app:a} for the full list of gates and coefficients).

\paragraph{Relaxation} We propose a relaxation of the WH transform-based decomposition that allows to differentiate w.r.t. its coefficients using the following three components:
\begin{enumerate*}[label=(\roman*)]
    \item Extend the component-wise basis transform $B$ to real-valued inputs, $\tilde B:[0,1]\to[-1,1]$, $\tilde B(x)=2x-1$,
    \item generalize the discrete WH coefficients $c_S$ to continuous, real-valued parameters $\tilde{c}_S$. This breaks bijectivity, creating a surjective mapping from the parameter space to all binary functions, as any truth table can now be represented by infinite combinations of generalized WH coefficients and
    \item Approximate the sign function with the sigmoid function in \cref{eq:walsh}, leading to $\tilde{f}(\mathbf x) = \sigma \left({l_{\mathbf c,\tilde B}(\mathbf x)}/{\tau}  \right)$, where $\tau$ is the \textit{temperature} parameter that controls the smoothness of the relaxation.
\end{enumerate*}
This decomposition provides a compact and differentiable parameterization with $2^n$ parameters, while still allowing to collapse back into one of the $2^{2^n}$ exact Boolean gates at inference time by choosing the closest truth table.

\paragraph{Reducing Discretization Gap}

To bridge the discretization gap between the continuous relaxation used during training and the hard decisions required during inference, we adopt Gumbel–Sigmoid reparameterization, a binary variant of the Gumbel–Softmax (Concrete) distribution, enabling gradient-based training with discrete variables~\cite{maddison2017concrete,jang2017categorical}.
Given a logit $l(x)$ and Gumbel noises $g_1,g_2\sim\text{Gumbel}(0,1)$, relaxed samples are computed as
$\tilde f_{\text{Gumbel}}(x)=\sigma\left((l(x) + g_1 - g_2)/\tau\right)$.
Discretization to a specific LUT can be obtained by deciding whether $\tilde f(x),\tilde f_{\text{Gumbel}}(x)<0.5$ which leads to a binary output. In our experiments, we also evaluated the performance of using this discretization already during training in the forward pass as a straight-through estimator.

This approach is motivated by two key observations.
First, injecting Gumbel noise during the forward pass induces stochastic smoothing, implicitly penalizing curvature and favoring flatter minima, which is known to improve generalization~\cite{yousefi2025mind}.
Second, unlike DLGNs, which optimize a surrogate objective misaligned with inference, Gumbel-based training mitigates this discrepancy.

\section{Experiments}
For our experiments, we restrict our study to the case of binary logic gates with two inputs, consistent with the setup considered in prior work. We evaluate model performance on the CIFAR-10 dataset, following standard pre-processing and an 80/20 split in training / validation data. The architectures of the models used are shown in~\cref{app:arch}.

In~\cref{fig:moneyplot}, we adopt the “large” model architecture introduced in the original DLGN paper~\cite{petersen2022deep} as the baseline. This MLP-style model contains 20,480,000 trainable parameters, while our WARP-LUT model based on the same architecture variant reduces this to 5,120,000 parameters. Both models were trained for 50,000 steps, albeit each WARP-LUT training step takes roughly one third of the wall clock time on an A100 GPU. After every 500 steps, we evaluate the models' performance on the validation set, both in the differentiable approximation ('Relaxed') used for training, as well as in its binarized mode ('Discrete'), where the most likely binary logic gate is plugged into each node. Overall, our implementation achieves a comparable final accuracy to the baseline while requiring only a quarter of the parameters and exhibiting significantly reduced resource utilization during training.

\begin{figure}
\centering
\includegraphics[width=0.9\textwidth]{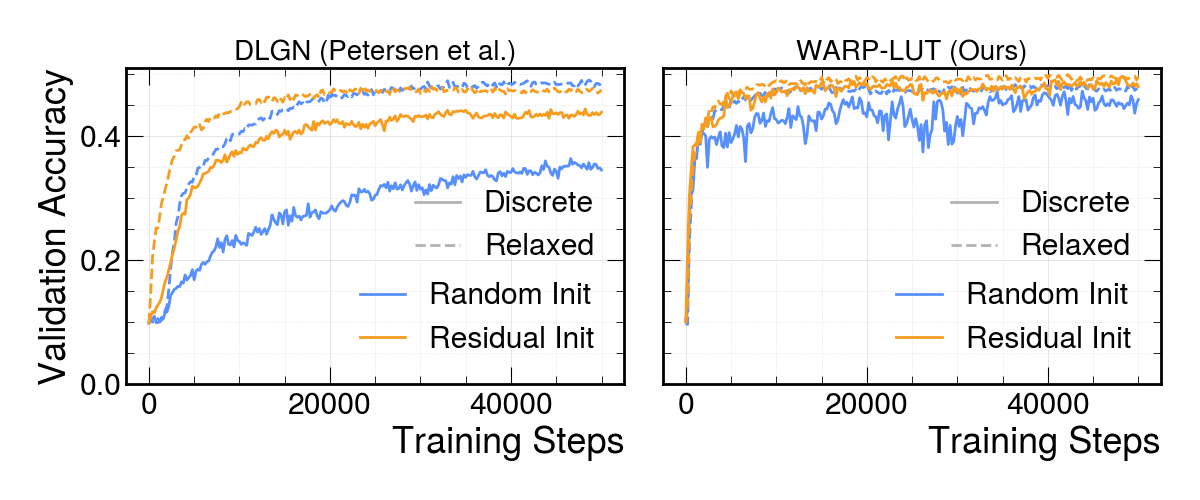}\\
\vspace*{-0.35cm}
\includegraphics[width=0.9\textwidth]{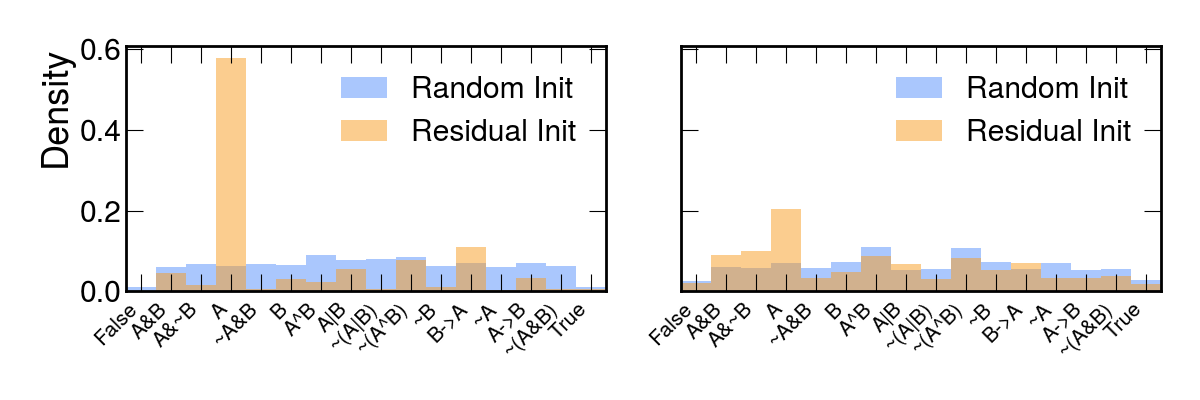}
\vspace*{-0.5cm}
\caption{Results of a very small (35968 gates) convolutional model on CIFAR-10 for DLGN (left) and WARP-LUT (right). We compare the validation accuracy, where the solid line shows the discrete mode and the dashed line shows the relaxed mode (top) along with the distribution of logic gates after training (bottom). Blue depicts random weight initialization, orange residual weight initialization. }
\label{fig:conv}
\end{figure}

In~\cref{fig:conv}, we show results for a very small convolutional model with only 575,488 and 143,872 trainable parameters in the DLGN-based and WARP-LUT implementation, respectively. We also compare random- and residual initialization of weights, where the latter gives stronger initial probability for the identity gate, which helps propagating gradients through many layers~\cite{petersen2024convolutional}. It can be seen that residual initialization helps both, DLGN and WARP-LUTs converge faster. In any case, WARP-LUTs converge in significantly fewer training steps than the baseline. Note that each training is also computationally less expensive (not factored into this plot). Unlike for WARP-LUTs, residual initialization for DLGNs also results in significantly more identity gates after the training, which should lead to a smaller model when deployed on hardware. However, this effect might disappear when running the training long enough for DLGN and WARP-LUT accuracy to match.


\section{Limitations and Future Work}

With access to larger computational resources, it would be valuable to investigate how WARP-LUTs perform for substantially larger models (on the order of one billion parameters) and across different architectures. All results in this paper were obtained using PyTorch’s default GPU support, without any dedicated CUDA kernels. Exploring optimized kernels for both DLGNs and WARP-LUTs will provide a more realistic assessment of their respective training efficiencies.

A key motivation for this work lies in scalability. While DLGNs scale doubly exponentially with the number of gate inputs, WARP-LUTs scale only exponentially, making it possible to learn four-input logic blocks with only 16 trainable parameters per node instead of roughly 65,000. If successful, this would open the door to even more efficient deployment on modern FPGAs, where multi-input look-up tables represent the fundamental computational units.
Our method will be compared to other LUT-based state-of-the-art methods, such as TreeLUT~\cite{khataei2025treelut} or Differentiable Weightless Neural Networks~\cite{bacellar2024differentiable}.



\section{Acknowledgements}
This work was supported by the National Science Foundation under Cooperative Agreements OAC-1836650 and PHY-2323298.

\printbibliography

@article{kunz1979equivalence,
  title={On the equivalence between one-dimensional discrete Walsh-Hadamard and multidimensional discrete Fourier transforms},
  author={Kunz},
  journal={IEEE Transactions on Computers},
  volume={100},
  number={3},
  pages={267--268},
  year={1979},
  publisher={IEEE}
}

@inproceedings{maddison2017concrete,
  title={The concrete distribution: A continuous relaxation of discrete random variables},
  author={Maddison, C and Mnih, A and Teh, Y},
  booktitle={Proceedings of the 5th International Conference on Learning Representations (ICLR)},
  year={2017},
  organization={International Conference on Learning Representations}
}

@inproceedings{jang2017categorical,
  title={Categorical Reparameterization with Gumbel-Softmax},
  author={Jang, Eric and Gu, Shixiang and Poole, Ben},
  booktitle={Proceedings of the 5th International Conference on Learning Representations (ICLR)},
  year={2017}
}

@inproceedings{aleksander2009brief,
  title={A brief introduction to weightless neural systems.},
  author={Aleksander, Igor and De Gregorio, Massimo and Fran{\c{c}}a, Felipe Maia Galvao and Lima, Priscila Machado Vieira and Morton, Helen},
  booktitle={Proceedings of the 17th European Symposium on Artificial Neural Networks (ESANN)},
  pages={299--305},
  year={2009}
}

@article{carneiro2019exact,
  title={The exact VC dimension of the WiSARD n-tuple classifier},
  author={Carneiro, Hugo CC and Pedreira, Carlos E and Fran{\c{c}}a, Felipe MG and Lima, Priscila MV},
  journal={Neural computation},
  volume={31},
  number={1},
  pages={176--207},
  year={2019},
  publisher={MIT Press}
}

@inproceedings{susskind2022weightless,
  title={Weightless neural networks for efficient edge inference},
  author={Susskind, Zachary and Arora, Aman and Miranda, Igor DS and Villon, Luis AQ and Katopodis, Rafael F and De Ara{\'u}jo, Leandro S and Dutra, Diego LC and Lima, Priscila MV and Fran{\c{c}}a, Felipe MG and Breternitz Jr, Mauricio and others},
  booktitle={Proceedings of the 31st International Conference on Parallel Architectures and Compilation Techniques (PACT)},
  pages={279--290},
  year={2022}
}

@article{susskind2023uleen,
  title={Uleen: A novel architecture for ultra-low-energy edge neural networks},
  author={Susskind, Zachary and Arora, Aman and Miranda, Igor DS and Bacellar, Alan TL and Villon, Luis AQ and Katopodis, Rafael F and De Ara{\'u}jo, Leandro S and Dutra, Diego LC and Lima, Priscila MV and Fran{\c{c}}a, Felipe MG and others},
  journal={Transactions on Architecture and Code Optimization (TACO)},
  volume={20},
  number={4},
  pages={1--24},
  year={2023},
  publisher={ACM}
}

@article{petersen2022deep,
  title={Deep differentiable logic gate networks},
  author={Petersen, Felix and Borgelt, Christian and Kuehne, Hilde and Deussen, Oliver},
  journal={Advances in Neural Information Processing Systems},
  volume={35},
  pages={2006--2018},
  year={2022}
}

@article{mommen2025method,
  title={A Method for Optimizing Connections in Differentiable Logic Gate Networks},
  author={Mommen, Wout and Keuninckx, Lars and Hartmann, Matthias and Wambacq, Piet},
  journal={arXiv preprint arXiv:2507.06173},
  year={2025}
}

@article{yue2024learning,
  title={Learning interpretable differentiable logic networks},
  author={Yue, Chang and Jha, Niraj K},
  journal={IEEE Transactions on Circuits and Systems for Artificial Intelligence},
  year={2024},
  publisher={IEEE}
}

@article{kresse2025scalable,
  title={Scalable Interconnect Learning in Boolean Networks},
  author={Kresse, Fabian and Yu, Emily and Lampert, Christoph H},
  journal={arXiv preprint arXiv:2507.02585},
  year={2025}
}

@article{kim2023deep,
  title={Deep stochastic logic gate networks},
  author={Kim, Youngsung},
  journal={IEEE Access},
  volume={11},
  pages={122488--122501},
  year={2023},
  publisher={IEEE}
}

@article{yousefi2025mind,
  title={Mind the Gap: Removing the Discretization Gap in Differentiable Logic Gate Networks},
  author={Yousefi, Shakir and Plesner, Andreas and Aczel, Till and Wattenhofer, Roger},
  journal={arXiv preprint arXiv:2506.07500},
  year={2025}
}

@inproceedings{bacellar2024differentiable,
  title={Differentiable weightless neural networks},
  author={Bacellar, Alan TL and Susskind, Zachary and Breternitz, Maur{\'\i}cio and John, Eugene and John, Lizy K and Lima, Priscila MV and Fran{\c{c}}a, Felipe MG and others},
  booktitle={Proceedings of the 41st International Conference on Machine Learning, PMLR},
  volume={235},
  pages={2277--2295},
  year={2024}
}

@article{petersen2024convolutional,
  title={Convolutional differentiable logic gate networks},
  author={Petersen, Felix and Kuehne, Hilde and Borgelt, Christian and Welzel, Julian and Ermon, Stefano},
  journal={Advances in Neural Information Processing Systems},
  volume={37},
  pages={121185--121203},
  year={2024}
}

@inproceedings{khataei2025treelut,
  title={TreeLUT: An Efficient Alternative to Deep Neural Networks for Inference Acceleration Using Gradient Boosted Decision Trees},
  author={Khataei, Alireza and Bazargan, Kia},
  booktitle={Proceedings of the 2025 ACM/SIGDA International Symposium on Field Programmable Gate Arrays (FPGA)},
  pages={14--24},
  year={2025}
}

@inproceedings{benamira2024truth,
  title={Truth table net: Scalable, compact \& verifiable neural networks with a dual convolutional small boolean circuit networks form},
  author={Benamira, Adrien and Peyrin, Thomas and Yap, Trevor and Gu{\'e}rand, Tristan and Hooi, Bryan},
  booktitle={Proceedings of the 33rd International Joint Conference on Artificial Intelligence (IJCAI)},
  year={2024}
}

@article{hubara2016binarized,
  title={Binarized neural networks},
  author={Hubara, Itay and Courbariaux, Matthieu and Soudry, Daniel and El-Yaniv, Ran and Bengio, Yoshua},
  journal={Advances in neural information processing systems},
  volume={29},
  year={2016}
}

@article{qin2020binary,
  title={Binary neural networks: A survey},
  author={Qin, Haotong and Gong, Ruihao and Liu, Xianglong and Bai, Xiao and Song, Jingkuan and Sebe, Nicu},
  journal={Pattern Recognition},
  volume={105},
  pages={107281},
  year={2020},
  publisher={Elsevier}
}

@article{nguyen2024bold,
  title={BOLD: Boolean Logic Deep Learning},
  author={Nguyen, Van Minh and Ocampo-Blandon, Cristian and Askri, Aymen and Leconte, Louis and Tran, Ba-Hien},
  journal={Advances in Neural Information Processing Systems},
  volume={37},
  pages={61912--61962},
  year={2024}
}

@inproceedings{chen2020addernet,
  title={AdderNet: Do we really need multiplications in deep learning?},
  author={Chen, Hanting and Wang, Yunhe and Xu, Chunjing and Shi, Boxin and Xu, Chao and Tian, Qi and Xu, Chang},
  booktitle={Proceedings of the 2020 IEEE/CVF Conference on Computer Vision and Pattern Recognition (CVPR)},
  pages={1468--1477},
  year={2020}
}

@inproceedings{elhoushi2021deepshift,
  title={Deepshift: Towards multiplication-less neural networks},
  author={Elhoushi, Mostafa and Chen, Zihao and Shafiq, Farhan and Tian, Ye Henry and Li, Joey Yiwei},
  booktitle={Proceedings of the 2021 IEEE/CVF Conference on Computer Vision and Pattern Recognition (CVPR)},
  pages={2359--2368},
  year={2021}
}

@article{sun2024gradient,
  title={Gradient-based automatic mixed precision quantization for neural networks on-chip},
  author={Sun, Chang and {\AA}rrestad, Thea K and Loncar, Vladimir and Ngadiuba, Jennifer and Spiropulu, Maria},
  journal={arXiv preprint arXiv:2405.00645},
  year={2024}
}

@incollection{gholami2022survey,
  title={A survey of quantization methods for efficient neural network inference},
  author={Gholami, Amir and Kim, Sehoon and Dong, Zhen and Yao, Zhewei and Mahoney, Michael W and Keutzer, Kurt},
  booktitle={Low-Power Computer Vision},
  pages={291--326},
  year={2022},
  publisher={Chapman and Hall/CRC}
}

@inproceedings{chmielneural,
  title={Neural gradients are near-lognormal: improved quantized and sparse training},
  author={Chmiel, Brian and Ben-Uri, Liad and Shkolnik, Moran and Hoffer, Elad and Banner, Ron and Soudry, Daniel},
  booktitle={Proceedings of the 9th International Conference on Learning Representations (ICLR)},
  year={2021}
}

@inproceedings{lin2018accelerating,
  title={Accelerating Convolutional Networks via Global \& Dynamic Filter Pruning.},
  author={Lin, Shaohui and Ji, Rongrong and Li, Yuchao and Wu, Yongjian and Huang, Feiyue and Zhang, Baochang},
  booktitle={Proceedings of the 27th International Joint Conference on Artificial Intelligence (IJCAI)},
  volume={2},
  number={7},
  pages={8},
  year={2018},
  organization={Stockholm}
}

@article{hoefler2021sparsity,
  title={Sparsity in deep learning: Pruning and growth for efficient inference and training in neural networks},
  author={Hoefler, Torsten and Alistarh, Dan and Ben-Nun, Tal and Dryden, Nikoli and Peste, Alexandra},
  journal={Journal of Machine Learning Research},
  volume={22},
  number={241},
  pages={1--124},
  year={2021}
}

@inproceedings{liu2022unreasonable,
  title={The Unreasonable Effectiveness of Random Pruning: Return of the Most Naive Baseline for Sparse Training},
  author={Liu, Shiwei and Chen, Tianlong and Chen, Xiaohan and Shen, Li and Mocanu, Decebal Constantin and Wang, Zhangyang and Pechenizkiy, Mykola},
  booktitle={Proceedings of the 10th International Conference on Learning Representations (ICLR)},
  year={2022},
  organization={OpenReview}
}

@article{sung2021training,
  title={Training neural networks with fixed sparse masks},
  author={Sung, Yi-Lin and Nair, Varun and Raffel, Colin A},
  journal={Advances in Neural Information Processing Systems},
  volume={34},
  pages={24193--24205},
  year={2021}
}

@article{lecun2015deep,
  title={Deep learning},
  author={LeCun, Yann and Bengio, Yoshua and Hinton, Geoffrey},
  journal={Nature},
  volume={521},
  number={7553},
  pages={436--444},
  year={2015},
  publisher={Nature Publishing Group UK London}
}

@article{Aarrestad_2021,
   title={Fast convolutional neural networks on FPGAs with hls4ml},
   volume={2},
   ISSN={2632-2153},
   url={http://dx.doi.org/10.1088/2632-2153/ac0ea1},
   DOI={10.1088/2632-2153/ac0ea1},
   number={4},
   journal={Machine Learning: Science and Technology},
   publisher={IOP Publishing},
   author={Aarrestad, Thea and Loncar, Vladimir and Ghielmetti, Nicolò and Pierini, Maurizio and Summers, Sioni and Ngadiuba, Jennifer and Petersson, Christoffer and Linander, Hampus and Iiyama, Yutaro and Di Guglielmo, Giuseppe and Duarte, Javier and Harris, Philip and Rankin, Dylan and Jindariani, Sergo and Pedro, Kevin and Tran, Nhan and Liu, Mia and Kreinar, Edward and Wu, Zhenbin and Hoang, Duc},
   year={2021},
   month=jul, pages={045015}
}

@misc{martins2025improvingearlydetectiongravitational,
      title={Improving early detection of gravitational waves from binary neutron stars using CNNs and FPGAs}, 
      author={Ana Martins and Melissa Lopez and Quirijn Meijer and Gregory Baltus and Marc van der Sluys and Chris Van Den Broeck and Sarah Caudill},
      year={2025},
      eprint={2409.05068},
      archivePrefix={arXiv},
      primaryClass={astro-ph.IM},
      url={https://arxiv.org/abs/2409.05068}, 
}

@misc{bhat2024machinelearningarbitrarysinglequbit,
      title={Machine Learning for Arbitrary Single-Qubit Rotations on an Embedded Device}, 
      author={Madhav Narayan Bhat and Marco Russo and Luca P. Carloni and Giuseppe Di Guglielmo and Farah Fahim and Andy C. Y. Li and Gabriel N. Perdue},
      year={2024},
      eprint={2411.13037},
      archivePrefix={arXiv},
      primaryClass={quant-ph},
      url={https://arxiv.org/abs/2411.13037}, 
}
\newpage

\appendix
\section{Source Code}
Our source code is available in GitHub: \href{https://github.com/ligerlac/torchlogix}{https://github.com/ligerlac/torchlogix}

\section{WH-Transform-based Decompositions for Two-Input Boolean Functions}
\label{app:a}
\begin{table}[h!]
\small
\caption{The 16 binary Boolean functions of two inputs, their truth tables (ordered as 00,01,10,11), and the corresponding WH coefficients, $(c_0,c_1,c_2,c_3)$.}
\centering
\renewcommand{\arraystretch}{1.2}
\begin{tabular}{c|c|c|c|c}
\hline
\# & Gate & Formula & Truth table & WH coeffs \\
\hline
0  & CONST0   & $0$                 & 0000 & $(-1,\,0,\,0,\,0)$ \\
1  & CONST1   & $1$                 & 1111 & $(+1,\,0,\,0,\,0)$ \\
2  & AND      & $a \land b$        & 0001 & $\left(-\tfrac{1}{2},\,\tfrac{1}{2},\,\tfrac{1}{2},\,\tfrac{1}{2}\right)$ \\
3  & OR       & $a \lor b$         & 0111 & $\left(\tfrac{1}{2},\,\tfrac{1}{2},\,\tfrac{1}{2},\,-\tfrac{1}{2}\right)$ \\
4  & XOR      & $a \oplus b$       & 0110 & $\left(0,\,0,\,0,\,-1\right)$ \\
5  & XNOR     & $\lnot(a \oplus b)$& 1001 & $\left(0,\,0,\,0,\,1\right)$ \\
6  & NAND     & $\lnot(a \land b)$ & 1110 & $\left(\tfrac{1}{2},\,-\tfrac{1}{2},\,-\tfrac{1}{2},\,-\tfrac{1}{2}\right)$ \\
7  & NOR      & $\lnot(a \lor b)$  & 1000 & $\left(-\tfrac{1}{2},\,-\tfrac{1}{2},\,-\tfrac{1}{2},\,\tfrac{1}{2}\right)$ \\
8  & $a \land \lnot b$ & $a \land \lnot b$   & 0010 & $\left(-\tfrac{1}{2},\,\tfrac{1}{2},\,-\tfrac{1}{2},\,-\tfrac{1}{2}\right)$ \\
9  & $\lnot a \land b$ & $\lnot a \land b$   & 0100 & $\left(-\tfrac{1}{2},\,-\tfrac{1}{2},\,\tfrac{1}{2},\,-\tfrac{1}{2}\right)$ \\
10 & ID(A)    & $a$                & 0011 & $(0,\,1,\,0,\,0)$ \\
11 & NOT(A)   & $\lnot a$          & 1100 & $(0,\,-1,\,0,\,0)$ \\
12 & ID(B)    & $b$                & 0101 & $(0,\,0,\,1,\,0)$ \\
13 & NOT(B)   & $\lnot b$          & 1010 & $(0,\,0,\,-1,\,0)$ \\
14 & IMP $(a \to b)$ & $\lnot a \lor b$ & 1101 & $\left(\tfrac{1}{2},\,-\tfrac{1}{2},\,\tfrac{1}{2},\,\tfrac{1}{2}\right)$ \\
15 & IMP $(b \to a)$ & $\lnot b \lor a$ & 1011 & $\left(\tfrac{1}{2},\,\tfrac{1}{2},\,-\tfrac{1}{2},\,\tfrac{1}{2}\right)$\\\hline
\end{tabular}
\label{tab:binary-gates}
\end{table}

\section{Model Architectures}
The architecture of the model used in \cref{fig:conv} is described in Table~\ref{tab:model-architecture}. The \texttt{ResidualLogicBlock} performs logic-based convolutional processing as described in \cite{petersen2024convolutional}. The only difference is the inclusion of a residual connection, which was found to increase training performance. 
Each \texttt{LogicDense} layer applies differentiable logic transformations as described in \cite{petersen2022deep}.   
The final \texttt{GroupSum} layer aggregates logical activations into class scores for CIFAR-10. 
\textbf{Model parameters:} For the \emph{Small} variant, $n_\text{bits}=3$, $k_\text{num}=32$, and $\tau=20$.
\label{app:arch}
\begin{table}[h!]
\caption{Architecture of the \textbf{CLGN CIFAR-10 Res} model.}
  \centering
  \small
  \label{tab:clgn_cifar10_smallres}
  \begin{tabular}{l c c c}
    \toprule
    Layer & Input dimension & Output dimension & Description \\
    \midrule
    ResidualLogicBlock &
    $3n_\text{bits} \times 32 \times 32$ &
    $2k_\text{num} \times 16 \times 16$ &
    Depth $=3$, receptive field $3\times3$, padding 1 \\
    Flatten &
    $2k_\text{num} \times 16 \times 16$ &
    $256 \times 2k_\text{num}$ &
    Spatial flattening of logic feature maps \\
    LogicDense$_1$ &
    $256 \times 2k_\text{num}$ &
    $512k_\text{num}$ &
    Fully connected logic layer \\
    LogicDense$_2$ &
    $512k_\text{num}$ &
    $256k_\text{num}$ &
    Fully connected logic layer \\
    LogicDense$_3$ &
    $256k_\text{num}$ &
    $320k_\text{num}$ &
    Fully connected logic layer \\
    GroupSum &
    $320k_\text{num}$ &
    $10$ &
    Grouped logic aggregation with temperature $\tau$ \\
    \bottomrule
  \end{tabular}
 \label{tab:model-architecture}
\end{table}

\end{document}